%% file: MAIN.tex
\documentclass{IEEEtran}
\IEEEoverridecommandlockouts

\makeatletter
\def\endthebibliography{%
  \def\@noitemerr{\@latex@warning{Empty `thebibliography' environment}}%
  \endlist
}
\makeatother

\usepackage[cmex10]{amsmath}
\usepackage[utf8]{inputenc} 
\usepackage[T1]{fontenc}    
\usepackage{hyperref}       
\usepackage{url}            
\usepackage{booktabs}       
\usepackage{amsfonts}       
\usepackage{nicefrac}       
\usepackage{microtype}      
\usepackage{xcolor}         
\usepackage{stfloats}
\usepackage{etoolbox}
\usepackage{cuted}
\usepackage{afterpage}
\usepackage{multirow}

\usepackage{amsmath,amssymb,amsthm,commath}
\usepackage{graphicx}
\usepackage{comment}
\usepackage{bbm}
\newtheorem{theorem}{Theorem}
\newtheorem{corollary}{Corollary}[theorem]

\newcommand*\diff{\mathop{}\!\mathrm{d}}
\makeatletter
\patchcmd{\@makecaption}
  {\scshape}
  {}
  {}
  {}
\makeatother

\begin{document}

\title{Copula Density Neural Estimation}

\author{%
    \IEEEauthorblockN{Nunzio A. Letizia\textsuperscript{\IEEEauthorrefmark{1}}, 
    Nicola Novello\textsuperscript{\IEEEauthorrefmark{1}}, and 
    Andrea M. Tonello \thanks{The authors are with the University of Klagenfurt - Institute of Networked and Embedded Systems. (e-mail: \{nunzio.letizia, nicola.novello, andrea.tonello\}@aau.at)}} 
}

\maketitle

\begingroup\renewcommand\thefootnote{\IEEEauthorrefmark{1}}
\footnotetext{Equal contribution}
\endgroup

\begin{abstract}
Probability density estimation from observed data constitutes a central task in statistics. 
In this brief, we focus on the problem of estimating the copula density associated to any observed data, as it fully describes the dependence between random variables. 
We separate univariate marginal distributions from the joint dependence structure in the data, the copula itself, and we model the latter with a neural network-based method referred to as copula density neural estimation (CODINE). Results show that the novel learning approach is capable of modeling complex distributions and can be applied for mutual information estimation and data generation.
\end{abstract}

\begin{IEEEkeywords}
copula density estimation, copula, deep learning, mutual information, data generation, $f$-divergence.
\end{IEEEkeywords}
\input{Introduction.tex}


\input{Theory.tex}


\input{Conclusions.tex}

\bibliographystyle{unsrt}
\bibliography{biblio}

\newpage 
\appendix

\input{Appendix.tex}

\end{document}

%% file: Introduction.tex
\section{Introduction}
\label{sec:introduction}
A natural way to discover data properties is to study the underlying probability density function (pdf). Parametric and nonparametric models \cite{Silverman86} are viable solutions for density estimation problems that deal with low-dimensional data. The former are typically used when a prior knowledge on the data structure (e.g. distribution family) is available. The latter, instead, are more flexible since they do not require any specification of the distribution's parameters. Practically, the majority of methods from both classes fail in estimating high-dimensional densities. Hence, some recent works leveraged deep neural networks as density estimators \cite{pmlr-v235-novello24a, PixelRNN,dinh2017density}.
Although significant efforts have been made to scale neural network architectures in order to improve their modeling capabilities, most of tasks translate into conditional distribution estimations. 
Instead, generative models attempt to learn the a-priori distribution to synthesize new data out of it. Deep generative models such as generative adversarial networks \cite{Goodfellow2014}, variational autoencoders \cite{Kingma2014} and diffusion models \cite{DDPMs}, tend to either implicitly estimate the underlying pdf or explicitly estimate a variational lower bound, providing the designer with no simple access to the investigated pdf.  
When the focus of the density estimation is to model the random vectors dependencies, it is possible to work with pseudo-observations, a projection of the collected observations into the uniform probability space via the probability integral transform (PIT) \cite{fisher1934statistical}. The probability density estimation becomes a copula density estimation problem. In this brief, we formulate and solve this problem using deep learning techniques. 
An intuitive approach can be to estimate the copula density with a neural network by minimizing the Kullback-Leibler (KL) divergence between the desired density and the output of the neural network. 
In the case of discrete random vectors, this approach corresponds, for instance, to the standard way of training neural networks for classification tasks, where the network's output distribution is ensured to be a probability mass function thanks to the usage of the softmax layer. However, when considering continuous random vectors, the network's output is not guaranteed to be a valid pdf and that may only be achieved by imposing additional constraints.
The envisioned copula density neural estimation method, referred to as CODINE, intrinsically guarantees that the estimated copula density is a valid pdf. CODINE is, to the best of our knowledge, the first neural estimator of nonparametric copula densities that can be learned with one neural network using one objective function. 
We present self-consistency tests and metrics that can be used to assess the quality of the estimator. In addition, we exploit the fact that the mutual information can be rewritten in terms of copula pdfs to estimate it using a slightly modified version of CODINE. Finally, we demonstrate the application of CODINE in the context of data generation.

The brief is organized as follows. Section \ref{sec:CE} introduces the copula and a brief description of related work. Section \ref{sec:codine} presents the copula density neural estimation approach as the solution of an optimization problem. Section \ref{sec:self} proposes self-consistency tests to assess the quality of the estimator. Section \ref{sec:applications} utilizes CODINE for copula density estimation, mutual information estimation and data generation. Finally, the conclusions are drawn.

%% file: Theory.tex
\section{Preliminaries}
\label{sec:CE}
\subsection{Copula}
Let us assume that the collected $n$ data observations $\{\mathbf{x}^{(1)},\dots,\mathbf{x}^{(n)}\}$ are produced by a fixed unknown or difficult to construct multivariate distribution of dimension $d$ with pdf $p_{X}(\mathbf{x}) = p_{X}(x_1, x_2, \dots, x_d)$ and cumulative distribution function (cdf) $F_{X}(\mathbf{x}) = P(X_1\leq x_1, \dots, X_d \leq x_d)$. 
Consider the univariate random variable $X_i$, whose marginal pdf $p_{X_i}(x_i)$ and cdf $F_{X_i}(x_i)$ are accessible since they can be obtained from the observations.
Then, the PIT is used to map the data into the uniform probability space, while the inverse transform sampling can be used to generate samples of $X$ given samples of a uniform distribution. In fact, if $U_i$ is a uniform random variable, then
$X_i = F^{-1}_{X_i}(U_i)$ is a random variable with cdf $F_{X_i}$.
Therefore, if the cdf is invertible, the transformation $u_i = F_{X_i}(x_i) \; \forall i=1,\dots,d$ projects the data $\mathbf{x}$ into the uniform probability space with finite distribution's support $u_i \in [0,1]$. The obtained transformed observations are typically called pseudo-observations.
In principle, the PIT is extremely beneficial: it offers a statistical normalization, thus a pre-processing operation that constitutes the first step of any deep learning pipeline.

To characterize the nature of the transformed data in the uniform probability space, it is convenient to introduce the concept of copula, a tool to analyze data dependence and construct multivariate distributions. Let $(U_1,U_2,\dots,U_d)$ be uniform random variables, then their joint cdf $F_{U}(\mathbf{u}) = P(U_1\leq u_1, \dots, U_d \leq u_d)$ is a copula $C:[0,1]^d \rightarrow [0,1]$ (see \cite{Nelsen2006}). The core of  copulas resides in Sklar's theorem \cite{Sklar} which states that if $F_{X}$ is a $d$-dimensional cdf with continuous marginals $F_{X_1},\dots,F_{X_d}$, then $F_{X}$ has a unique copula representation
\begin{equation}
\label{SklarF}
F_{X}(x_1,\dots,x_d) = C_{U}(F_{X_1}(x_1),\dots,F_{X_d}(x_d)).
\end{equation}
Moreover, when the multivariate distribution is described in terms of the pdf $p_{X}$, it holds that
\begin{equation}
\small
\label{Sklarf}
p_{X}(x_1,\dots,x_d) = c_{U}(F_{X_1}(x_1),\dots,F_{X_d}(x_d))\cdot \prod_{i=1}^{d}{p_{X_i}(x_i)},
\end{equation}
where $c_{U}$ is the density of the copula.

The relation in \eqref{Sklarf} is the fundamental building block of this paper. It separates the dependence internal structure of $p_{X}$ into two distinct components: the product of all the marginals $p_{X_i}$ and the density of the copula $c_{U}$. By nature, the former accounts only for the marginal information, thus, the statistics of each univariate variable. The latter, instead, accounts only for the joint dependence of data.

Considering the fact that building the marginals is usually a straightforward task, the estimation of the empirical joint density $\hat{p}_{X}(\mathbf{x})$ of the observations $\{\mathbf{x}^{(1)},\dots,\mathbf{x}^{(n)}\}$ passes through the estimation of the empirical copula density $\hat{c}_{U}(\mathbf{u})$ of the pseudo-observations $\{\mathbf{u}^{(1)},\dots,\mathbf{u}^{(n)}\}$.   

\subsection{Related Work}
To study the copula structure, it is possible to build 
a simple nonparametric estimator of the cdf. For a finite set of samples, a naive copula estimator has expression
\begin{equation}
\hat{C}_n(\mathbf{u}) = \frac{1}{n}\sum_{j=1}^{n}{\mathbbm{1}_{\{U_{1}^{(j)} < u_1, \dots, U_{d}^{(j)} < u_d\}}}
\end{equation}
where $n$ is the number of observations, $U_{i}^{(j)}$ denotes the $j$-th pseudo-observation of the $i$-th random variable, with $i=1,\dots,d$, and $\mathbbm{1}_A$ is the indicator function. 
However, strong complexity limitations occur for increasing values of $d$, forcing to either move towards parametric families of multivariate copulas or towards learning-based approaches. The latter have more flexibility as they do not pose any assumptions on the data distribution. For the former, common models are the multivariate Gaussian copula with correlation matrix $\Sigma$ or the multivariate Student's t-copula with $\nu$ degrees of freedom and correlation matrix $\Sigma$, suitable for extreme value dependence \cite{t-copula}. 
Archimedean copulas assume the form $C(\mathbf{u})=\phi^{-1}(\phi(u_1)+\dots+\phi(u_d))$ where $\phi$ is the generator function of the Archimedean copula. Such structure is said to be exchangeable, i.e., the components can be swapped indifferently, but its symmetry introduces modeling limitations. Multivariate copulas built using bivariate pair-copulas, also referred to as vine copulas, represent a more flexible model but the selection of the vine tree structure and the pair copula families is a complex task \cite{aas2009pair, nagler2016evading, Czado2021}. In particular, there are three types of vine copulas: regular vine (R-vine), drawable vine (D-vine), and canonical vine (C-vine) \cite{kurowicka2006uncertainty}. Other density estimation techniques relying on parametric copulas have been proposed in \cite{scott2015multivariate, bayestehtashk2013parsimonious}. 
Copula Bayesian Networks \cite{elidan2010copula} model multivariate distributions based on a directed graph representation, merging the copula framework with Bayesian networks. However, they require the choice for an appropriate local copula function for each conditional distribution. In \cite{majdara2019nonparametric}, the authors propose an algorithm combining diffusion-based kernel density estimation and Bayesian sequential partitioning. Deep Archimedean Copulas \cite{ling2020deep} learns the generator of an Archimedean copula using a neural network. 

To our knowledge, there is absence of neural copula density estimators able to learn nonparametric copula densities with one simple fully connected or convolutional network and one objective function. The work that gets closer in solving this challenge is the one proposed in \cite{zeng2022neural}. However, it estimates the cdf of the copula by requiring $d+1$ neural networks trained with at least four different cost functions to impose constraints that guarantee the estimate of a valid copula, thus leading to an extremely computationally-intensive framework. 

Besides density estimation, in the machine learning literature the usage of copulas ranged various applications \cite{elidan2013copulas}, such as economical market modeling \cite{xu2023copula}, transfer learning \cite{ma2023deep}, and imitation learning \cite{wang2021multi}. 
In the last years, copulas have also been used as generative approaches. 
In \cite{tagasovska2019copulas}, the authors propose to plug a vine copula into an autoencoder to obtain a generative model. 
In \cite{Letizia2020}, the authors propose a modified version of GANs leveraging the copula transformation. 
Implicit generative copulas \cite{janke2021implicit} generate data starting from a Gaussian distribution and then fit the data to the dataset copula distribution. 

\section{Copula Density Neural Estimation}
\label{sec:codine}

In the following, we propose to use deep neural networks to model dependencies in high-dimensional data, and in particular to estimate the copula pdf. The proposed framework relies on the following simple idea: we can measure the statistical distance  between the pseudo-observations and uniform i.i.d. realizations using neural network parameterization. Surprisingly, by maximizing a variational lower bound on a divergence measure, we obtain the copula density neural estimator.

\subsection{Variational Formulation}
\label{subsec:f-div}
The $f$-divergence $D_f(P||Q)$ is a measure of dependence between two distributions $P$ and $Q$. In detail, let $P$ and $Q$ be two probability measures on $\mathcal{X}$ and assume they possess densities $p$ and $q$, then the $f$-divergence is defined as follows
\begin{equation}
D_f(P||Q) = \int_{\mathcal{X}}{q(x)f\biggl(\frac{p(x)}{q(x)}\biggr)\diff x},
\end{equation}
where $\mathcal{X}$ is a compact domain and the function $f:\mathbb{R}_+ \to \mathbb{R}$ is convex, lower semicontinuous and satisfies $f(1)=0$. Let $f^*$ be the \textit{Fenchel conjugate} of $f$, defined as $f^*(t) = \sup_{u \in dom_f} \left\{ ut - f(u) \right\}$,
with $dom_f$ being the domain of the function $f$.

One might be interested in studying the particular case when the two densities $p$ and $q$ correspond to $c_U$ and $\pi_U$, respectively, where $\pi_U$ describes a multivariate uniform distribution on $[0,1]^d$. In such situation, it is possible to obtain a copula density expression via the variational representation of the $f$-divergence. The following Theorem formulates an optimization problem whose solution yields to the desired copula density.
 
\begin{theorem}
\label{theorem:Theorem1}
Let $\mathbf{u} \sim c_U(\mathbf{u})$ be $d$-dimensional samples drawn from the copula density $c_U$. Let $f^*$ be the Fenchel conjugate of $f:\mathbb{R}_+ \to \mathbb{R}$, a convex lower semicontinuous function that satisfies $f(1)=0$ and has derivative $f^{\prime}$. If $\pi_U(\mathbf{u})$ is a multivariate uniform distribution with i.i.d. components on the unit cube $[0,1]^d$ and $\mathcal{J}_{f}(T)$ is a value function defined as 
\begin{equation}
\mathcal{J}_{f}(T) = \mathbb{E}_{\mathbf{u} \sim c_{U}(\mathbf{u})}\biggl[T\bigl(\mathbf{u}\bigr)\biggr] -\mathbb{E}_{\mathbf{u} \sim \pi_{U}(\mathbf{u})}\biggl[f^*\biggl(T\bigl(\mathbf{u}\bigr)\biggr)\biggr],
\label{eq:discriminator_function_f}
\end{equation}
then
\begin{equation}
\label{eq:optimal_ratio_T}
c_U(\mathbf{u}) = \bigl(f^{*}\bigr)^{\prime} \bigl(\hat{T}(\mathbf{u})\bigr), 
\end{equation}
where
\begin{equation}
\hat{T}(\mathbf{u}) = \arg \max_T \mathcal{J}_f(T).
\end{equation}
\end{theorem}

The proof of Theorem \ref{theorem:Theorem1} is reported in Appendix \ref{subsec:proof_thm1}. 
Notice that the density of the copula can be derived with the same approach also by working in the sample domain. Indeed, when $p$ and $q$ correspond to the joint and the product of the marginals, respectively, the following corollary holds. Examples of generator functions $f$ are given in Tab.~\ref{tab:generators}.

\begin{corollary}{}{}
\label{cor:cor1}
Let $\mathbf{x} \sim p_X(\mathbf{x})$ be $d$-dimensional samples drawn from the joint density $p_X$. Let $f^*$ be the Fenchel conjugate of $f:\mathbb{R}_+ \to \mathbb{R}$, a convex lower semicontinuous function that satisfies $f(1)=0$ and has derivative $f^{\prime}$. If $\pi_X(\mathbf{x})$ is the product of the marginals $p_{X_i}(x_i)$ and $\mathcal{J}_{f}(T)$ is a value function defined as 
\begin{equation}
\label{eq:codine_discriminative_joint_marg}
\mathcal{J}_{f,x}(T) = \mathbb{E}_{\mathbf{x} \sim p_{X}(\mathbf{x})}\biggl[T\bigl(\mathbf{x}\bigr)\biggr] -\mathbb{E}_{\mathbf{x} \sim \pi_{X}(\mathbf{x})}\biggl[f^*\biggl(T\bigl(\mathbf{x})\bigr)\biggr)\biggr],
\end{equation}
then
\begin{equation}
c_U(\mathbf{u}) = \bigl(f^{*}\bigr)^{\prime} \bigl(\hat{T}(\mathbf{F}_X^{-1}(\mathbf{u}))\bigr)
\end{equation}
is the copula density, where
\begin{equation}
\mathbf{F}_X^{-1}(\mathbf{u}) := [F_{X_i}^{-1}(u_i),\dots,F_{X_d}^{-1}(u_d)]
\end{equation}
and
\begin{equation}
\hat{T}(\mathbf{x}) = \arg \max_T \mathcal{J}_{f,x}(T).
\end{equation}
\end{corollary}
The proof of Corollary \ref{cor:cor1} is reported in Appendix \ref{subsec:proof_cor1}. 
A great advantage of the formulation in \eqref{eq:discriminator_function_f} comes from the second expectation term. Conversely to the variational discriminative formulation in \eqref{eq:codine_discriminative_joint_marg} that tests jointly with marginals samples, the comparison in \eqref{eq:discriminator_function_f} is made between samples from the joint copula structure and independent uniforms. The latter can be easily generated without the need of any scrambler that factorizes $p_X$ into the product of the marginal pdfs. More precisely, a derangement type of shuffling mechanism would be required as proved in \cite{letizia2024Mutual}. On the other hand, \eqref{eq:discriminator_function_f} needs samples from the copula, thus, needs an estimate of the marginals of $X$ to apply the PIT. 

\subsection{Parametric Implementation}
\label{subsec:implementation}
To proceed, we propose to parametrize $T(\mathbf{u})$ with a deep neural network $T_{\theta}$ of parameters $\theta$ and maximize $\mathcal{J}_{f}(T)$ with gradient ascent and back-propagation 
\begin{equation}
    \hat{\theta} = \arg \max_{\theta} \mathcal{J}_f(T_{\theta}).
\end{equation}

\begin{table}
\caption{List of generator and conjugate functions used in the experiments.}
\begin{tabular}{l|l|l}
Name & Generator $f(u)$                   & Conjugate $f^*(t)$  \\ \hline
GAN  & $u\log u -(u+1)\log (u+1)+\log(4)$ & $-\log (1-\exp(t))$ \\ \hline
KL   & $u\log u$                          & $\exp(t-1)$         \\ \hline
HD   & $(\sqrt{u}-1)^2$                   & $t/(1-t)$     \\ \hline
\end{tabular}
\label{tab:generators}
\end{table} 

The resulting estimator of the copula density reads as follows
\begin{equation}
\hat{c}_U(\mathbf{u}) = \bigl(f^{*}\bigr)^{\prime} \bigl(T_{\hat{\theta}}(\mathbf{u})\bigr),
\end{equation}
and its training procedure enjoys two normalization properties. 
The former consists in a natural normalization of the input data in the interval $[0,1)$ via PIT that facilitates the training convergence and helps producing improved dependence measures \cite{Poczos2012}. The latter normalization property is perhaps at the core of the proposed methodology. The typical problem in creating neural density estimators is to enforce the network to return densities that integrate to one
\begin{equation}
\label{eq:density_test}
\int_{\mathbb{R}^d}{p_X(\mathbf{x};\theta)\diff \mathbf{x}} = 1
\end{equation}
Energy-based models have been proposed to tackle such constraint, but they often produce intractable densities (due to the normalization factor, see \cite{Papamakarios2015}). Normalizing flows \cite{Rezende2015} provide exact likelihoods but they are limited in representation. In contrast, the discriminative formulation of \eqref{eq:discriminator_function_f} produces a copula density neural estimator that naturally favors a solution of \eqref{eq:density_test}, without any architectural modification or regularization term.

\section{Self-Consistency Tests}
\label{sec:self}
When the copula density is known, it is possible to assess the quality of the copula density neural estimator $\hat{c}_{U}(\mathbf{u})$ by computing the KL divergence between the true and the estimated copulas
\begin{equation}
Q_c = D_{\text{KL}}(c_{U}||\hat{c}_{U}) = \mathbb{E}_{\mathbf{u}\sim c_{U}(\mathbf{u})}\biggl[\log\frac{c_{U}(\mathbf{u})}{\hat{c}_{U}(\mathbf{u})}\biggr].
\end{equation}

Once the dependence structure is characterized via a valid copula density $\hat{c}_{U}(\mathbf{u})$, a multiplication with the estimated marginal components $\hat{p}_{X_i}(x_i)$, $\forall i=\{1,\dots,d\}$ yields the estimate of the joint pdf $\hat{p}_{X}(\mathbf{x})$. In general, it is rather simple to build one-dimensional marginal density estimates $\hat{p}_{X_i}(x_i)$, e.g., using histograms or kernel functions. \\

To assess the quality of the copula density estimator $\hat{c}_U(\mathbf{u})$ when there is no ground-truth, we propose the following set of self-consistency tests over the basic property illustrated in \eqref{eq:density_test}. In particular,
\begin{enumerate}
\item if $\hat{c}_U(\mathbf{u})$ is a well-defined density and $\hat{c}_U(\mathbf{u})=c_U(\mathbf{u})$, then the following relation must hold
\begin{equation}
\mathbb{E}_{\mathbf{u} \sim \pi_{U}(\mathbf{u})}\bigl[\hat{c}_U(\mathbf{u})\bigr]=1,
\end{equation}
\item in general, for any $n$-th order moment, if $\hat{c}_U(\mathbf{u})$ is a well-defined density and $\hat{c}_U(\mathbf{u})=c_U(\mathbf{u})$, then
\begin{equation}
\mathbb{E}_{\mathbf{u} \sim \pi_{U}(\mathbf{u})}\bigl[\mathbf{u}^n\cdot \hat{c}_U(\mathbf{u})\bigr]=\mathbb{E}_{\mathbf{u} \sim c_{U}(\mathbf{u})}\bigl[\mathbf{u}^n\bigr].
\end{equation}
\end{enumerate}
The first test verifies that the copula density integrates to one while the second set of tests extends the first test to the moments of any order. Similarly, joint consistency tests can be defined, e.g., the Spearman rank correlation $\rho_{X,Y}$ \cite{spearman1961proof} between pairs of variables can be rewritten in terms of their joint copula density $\hat{c}_{UV}$ and it reads as follows
\begin{equation}
\rho_{X,Y} = 12 \cdot\mathbb{E}_{(\mathbf{u},\mathbf{v}) \sim \pi_{U}(\mathbf{u})\pi_{V}(\mathbf{v})}\bigl[\mathbf{u} \mathbf{v}\cdot \hat{c}_{UV}(\mathbf{u},\mathbf{v})\bigr]-3.
\end{equation}

\section{Results}
\label{sec:applications}

In this section, we first provide the details of the code implementation, and then we present the results attained by CODINE for density estimation, mutual information estimation, and data generation. A summary of the objective functions and tasks that CODINE tackles is reported in Tab.~\ref{tab:summary_codine}. 
\begin{table*}
	\centering
	\caption{Summary of the possible usages of CODINE.}
    \resizebox{\textwidth}{!}{%
	\begin{tabular}{ c  c  c  c  c } 
        \toprule
        \multicolumn{5}{c}{\textbf{CODINE}} \\
        \toprule
		   &\textbf{Density estimation} & \textbf{MI estimation} & \multicolumn{2}{c}{\textbf{Data generation}} \\
         & & & \textbf{Gibbs} & \textbf{GAN} \\
        \midrule 
		  \textbf{Objective function} & $\mathcal{J}_f(T_\theta)$ \eqref{eq:discriminator_function_f} & $\mathcal{J}_{f,\text{MI}}(T_\theta)$ \eqref{eq:fDIME_copula} & $\mathcal{J}_f(T_\theta)$ \eqref{eq:discriminator_function_f} & $\mathcal{J}_f(T_\theta)$ \eqref{eq:discriminator_function_f} $, \> \mathcal{J}_{\text{MMD}}(G_{\theta_G})$ \eqref{eq:MMD-metric}  \\
        \textbf{Task solution} & $\bigl(f^{*}\bigr)^{\prime} \bigl(T_{\hat{\theta}}(\mathbf{u})\bigr)$ & $\mathbb{E}_{(\mathbf{u},\mathbf{v}) \sim c_{UV}(\mathbf{u},\mathbf{v})}\bigl[ \log \bigl(\bigl(f^{*}\bigr)^{\prime}\bigl(T_{\hat{\theta}}(\mathbf{u},\mathbf{v})\bigr) \bigr) \bigr]$ & Gibbs$\bigl(\bigl(f^{*}\bigr)^{\prime} \bigl(T_{\hat{\theta}}(\mathbf{u})\bigr)\bigr)$ & $G_{\hat{\theta}_G}(\mathbf{v})$ \\
		\midrule
	\end{tabular}
    }
	\label{tab:summary_codine}
\end{table*}

\subsection{Implementation Details}

For the experiments on density estimation and toy dataset generation, we use a small fully connected neural network with 2 hidden layers comprising 100 neurons. For mutual information estimation, we use a small fully connected neural network with 2 hidden layers comprising 256 neurons. For image generation, the architecture used is represented in Fig.~\ref{fig:codine_architecture}, where the copula is estimated by a fully connected neural network with 2 hidden layers having 128 and 50 neurons, and the autoencoder comprises two 6-layer CNNs for encoding and decoding. Optimization is executed using Adam \cite{kingma2014adam}.
Since at convergence the network outputs a transformation of the copula density evaluated at the input $\mathbf{u}$, the final layer possesses a unique neuron with activation function that depends on the generator $f$ (see the code\footnote{\url{https://github.com/tonellolab/CODINE-copula-estimator}} for more details).

\subsection{Copula Density Estimation}
\label{sec:metrics}
\begin{figure}[t!]
	\centering
	\includegraphics[scale=0.2]{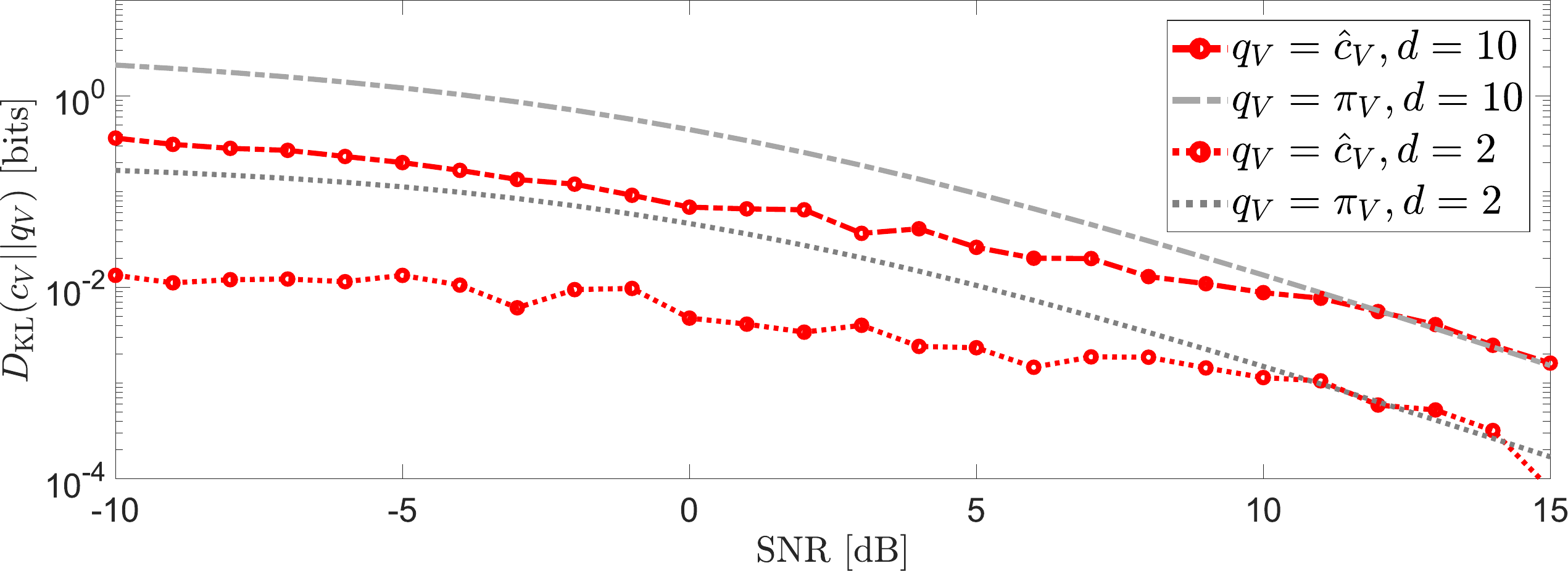}
	\caption{KL divergence between estimated and true copula in an ACGN channel as a function of the signal-to-noise ratio (SNR) and for different dimensionality $d$ of the input. The comparison with a flat copula density $\pi_V$ is also reported (gray curves).}
	\label{fig:q_c}
\end{figure}
\begin{figure}[t!]
	\centering
	\includegraphics[scale=0.225]{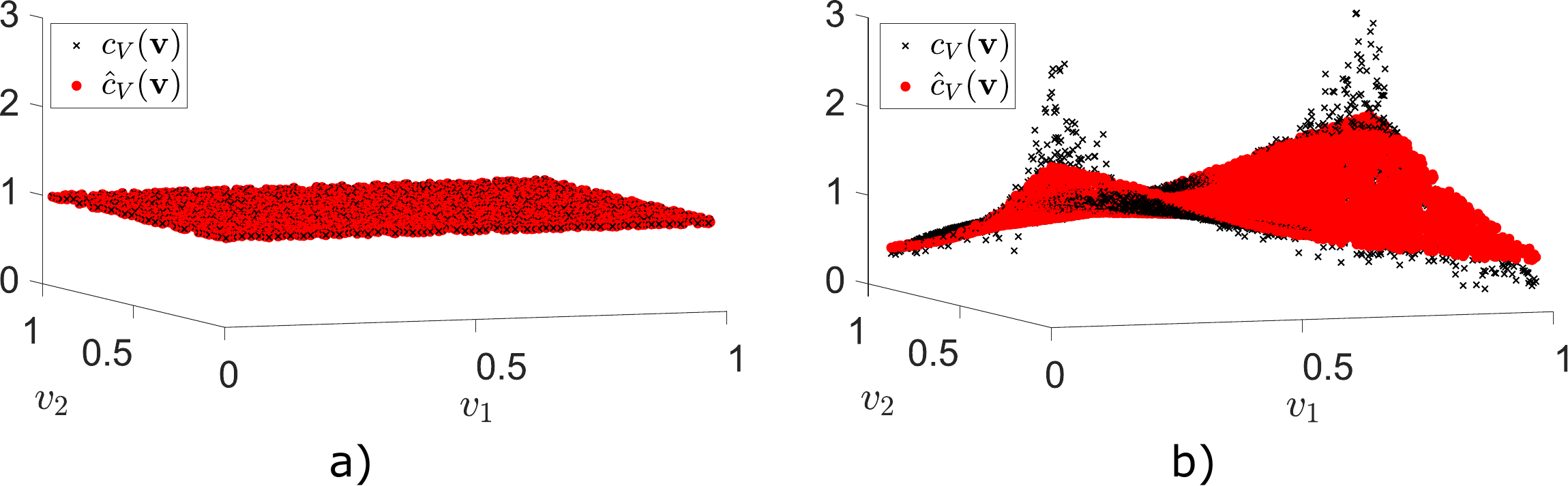}
	\caption{Ground-truth and estimated copula density (SNR=$0$ dB) at the channel output ($d=2$) using the GAN $f$ generator for: a) uncorrelated noise $\rho=0$. b) correlated noise with coefficient $\rho=0.5$.}
	\label{fig:copula_approx}
\end{figure}
Now, as a first example to validate the density estimator, we consider the transmission of $d$-dimensional Gaussian samples over an additive colored Gaussian channel (ACGN). Given the ACGN model $Y=X+N$, where $X \sim \mathcal{N}(0,\mathbb{I})$ and $N \sim \mathcal{N}(0,\Sigma_N)$, it is simple to obtain closed-form expressions for the probability densities involved. In particular, the copula density of the output $Y$ reads as in \eqref{eq:copula_gaussian_y}, where $\mathbf{F}_X(\mathbf{x})$ is an operator that element-wise applies the PIT (via Gaussian cumulative distributions) to the components of $\mathbf{x}$ such that $(\mathbf{u},\mathbf{v}) = (\mathbf{F}_X(\mathbf{x}),\mathbf{F}_Y(\mathbf{y}))$ and $\tilde{\Sigma}_N = \Sigma_N \odot \mathbb{I}$, where $A \odot B$ denotes the Hadamard product between $A$ and $B$.

\begin{align}
\label{eq:copula_gaussian_y}
c_{V}(\mathbf{v}) = \; & \sqrt{\frac{\det(\tilde{\Sigma}_N+\mathbb{I})}{\det(\Sigma_N+\mathbb{I})}}{\exp\biggl(-\frac{1}{2}\bigl(\mathbf{F}_Y^{-1}(\mathbf{v})\bigr)^T\bigl(\bigl(\Sigma_N+\mathbb{I}\bigr)^{-1}} \nonumber \\
&-{ \bigl(\tilde{\Sigma}_N+\mathbb{I}\bigr)^{-1}\bigr)\bigl(\mathbf{F}_Y^{-1}(\mathbf{v})\bigr)\biggr)}.
\end{align}

In Fig.~\ref{fig:q_c}, we illustrate the KL divergence (in bits) between the ground-truth and the neural estimator obtained using the GAN generator function reported in Tab.~\ref{tab:generators}. To work with non-uniform copula structures, we study the case of a non-diagonal noise covariance matrix $\Sigma_N$. In particular, we impose a tridiagonal covariance matrix such that $\Sigma_N=\sigma_N^2 R$ where $R_{i,i} = 1$ with $i=1,\dots,d$, and $R_{i,i+1}=\rho$, with $i=1,\dots,d-1$ and $\rho=0.5$. Moreover, Fig.~\ref{fig:q_c} also depicts the quality of the approximation for different values of the signal-to-noise ratio (SNR), defined as the reciprocal of the noise power $\sigma_N^2$, and for different dimensions $d$. To provide a numerical comparison, we also report the KL divergence $D_{\text{KL}}(c_{V}||\pi_{V})$ between the ground-truth and the flat copula density $\pi_V = 1$. It can be shown that when $c_V$ is Gaussian, we obtain
\begin{equation}
D_{\text{KL}}(c_{V}||\pi_{V}) = \frac{1}{2}\log\biggl(\frac{\det(\tilde{\Sigma}_N+\mathbb{I})}{\det(\Sigma_N+\mathbb{I})}\biggr).
\end{equation}

Notice that in Fig.~\ref{fig:q_c} we use the same simple neural network architecture for both $d=2$ and $d=10$. Nonetheless, CODINE can accurately approximate multidimensional densities even without any further hyper-parameter search. 
Fig.~\ref{fig:copula_approx}a reports a comparison between ground-truth and estimated copula densities at $0$ dB in the case of independent components ($\rho=0$) and correlated components ($\rho=0.5$). It is worth mentioning that when there is independence between components, the copula density is everywhere unitary $c_V(\mathbf{v})=1$. Hence, independence tests can be derived based on the structure of the estimated copula via CODINE, but we leave it for future discussions.

In addition, we test CODINE in the more challenging scenario of estimating multimodal distributions. In particular, we estimate the copula of a mixture of Gaussians (MoG) and compare the estimates obtained by CODINE, Gaussian copula (where the covariance matrix of the data is estimated using maximum likelihood), and C-vine copula, with the estimate obtained from the pseudo-observations using Kernel Density Estimation (KDE), in Fig.~\ref{fig:MoG}. CODINE's estimate is visibly close to the KDE estimate. Differently, the Gaussian and Vine copulas obtain significantly different estimates. Fig.~\ref{fig:MoG} demonstrates the effectiveness of CODINE in estimating complex distributions, where alternative copula models fail. \\ 
\begin{figure*}[t!]
	\centering
	\includegraphics[width=\textwidth]{pics/total_copula_comparison_KL_3_components.png}
	\caption{Estimate of the copula density of a Mixture of Gaussians. From left to right: samples from the MoG, copula density fitted from the pseudo-observations using KDE, CODINE copula density, Gaussian copula density, c-vine copula density. CODINE's estimate is obtained using the generator function of the KL divergence.}
	\label{fig:MoG}
\end{figure*}
Computationally, the proposed neural copula density estimation task requires an additional $O(d\cdot N)$ w.r.t. a standard neural network training for each epoch, which is necessary to perform the PIT using the empirical cumulative distribution functions, where $N$ is the batch size. 

\subsection{Mutual Information Estimation}
Given two random variables, $X$ and $Y$, the mutual information $I(X;Y)$ quantifies the statistical dependence between $X$ and $Y$. It measures the amount of information obtained about one variable via the observation of the other and it can be rewritten also in terms of KL divergence as
${I}(X;Y) = D_{\text{KL}}(p_{XY}||p_Xp_Y)$.
From Sklar's theorem, it is simple to show that the mutual information can be computed using only copula densities as follows
\begin{equation}
\label{eq:mutual_information_copula}
{I}(X;Y) = \mathbb{E}_{(\mathbf{u},\mathbf{v})\sim c_{UV}(\mathbf{u},\mathbf{v})}\biggl[\log\frac{c_{UV}(\mathbf{u},\mathbf{v})}{c_U(\mathbf{u}) c_V(\mathbf{v})}\biggr],
\end{equation}
where $c_{UV}$ is the copula density associated to the pseudo-observations of $X$ and $Y$.

\begin{figure}[t]
	\centering
	\includegraphics[scale=0.21]{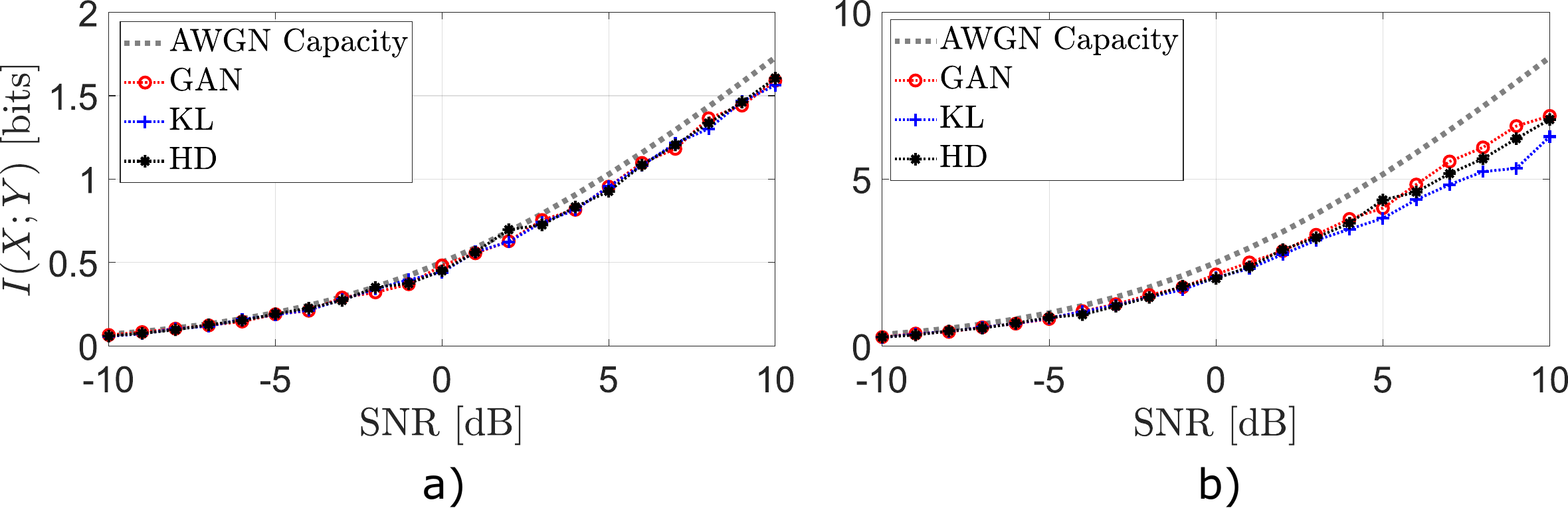}
	\caption{Estimated mutual information $I(X;Y)$ via joint copula $c_{UV}$ with different generators $f$ for: a) $d=1$. b) $d=5$.}
	\label{fig:i_x_y}
\end{figure}

Therefore, \eqref{eq:mutual_information_copula} requires three separate copula densities estimators, each of which is obtained as explained in Section \ref{sec:codine}. Alternatively, one could learn the copulas density ratio via maximization of the variational lower bound on the mutual information. Using again Fenchel duality, the KL divergence
\begin{equation}
\small
D_{\text{KL}}(c_{UV}||c_Uc_V) = \int_{[0,1]^{2d}}{c_{UV}(\mathbf{u},\mathbf{v})\log\biggl(\frac{c_{UV}(\mathbf{u},\mathbf{v})}{c_U(\mathbf{u})c_V(\mathbf{v})}\biggr)\diff \mathbf{u} \diff \mathbf{v}}
\end{equation}
corresponds to the supremum over $T$ (referred to as $T_{\hat{\theta}}$) of
\begin{equation}
\label{eq:DIME_copula}
\small
\mathcal{J}_{\text{KL}}(T_\theta) = \mathbb{E}_{c_{UV}}\biggl[T_\theta\bigl(\mathbf{u},\mathbf{v}\bigr)\biggr] - \mathbb{E}_{c_{U}c_{V}}\biggl[\exp{\bigl(T_\theta \bigl(\mathbf{u},\mathbf{v }\bigr)-1 \bigr)}\biggr].
\end{equation}

When $X$ is a univariate random variable, its copula density $c_U$ is unitary. Notice that \eqref{eq:DIME_copula} can be seen as a special case of the more general \eqref{eq:discriminator_function_f} when $f$ is the generator of the KL divergence and the second expectation is not done over independent uniforms with distribution $\pi_U$ but over samples from the product of copula densities $c_U\cdot c_V$
\begin{equation}
\label{eq:fDIME_copula}
\small
\mathcal{J}_{f,\text{MI}}(T_\theta) = \mathbb{E}_{c_{UV}}\biggl[T_\theta\bigl(\mathbf{u},\mathbf{v}\bigr)\biggr] - \mathbb{E}_{c_{U}c_{V}}\biggl[f^*{\bigl(T_\theta \bigl(\mathbf{u},\mathbf{v }\bigr)\bigr)}\biggr].
\end{equation}
Furthermore, similarly to what was shown in \cite{letizia2024Mutual}, it is possible to obtain low variance mutual information estimates with any $f$-divergence by extracting the estimated densities ratio and plugging it in the MI definition as
\begin{align}
    \hat{I}(X;Y) = \mathbb{E}_{(\mathbf{u},\mathbf{v}) \sim c_{UV}(\mathbf{u},\mathbf{v})}\biggl[ \log \biggl(\bigl(f^{*}\bigr)^{\prime}\bigl(T_{\hat{\theta}}(\mathbf{u},\mathbf{v})\bigr) \biggr) \biggr],
\end{align}
where $\bigl(f^{*}\bigr)^{\prime}\bigl(T_{\hat{\theta}}(\mathbf{u},\mathbf{v})\bigr) = c_{UV}(\mathbf{u}, \mathbf{v})/c_U(\mathbf{u})c_V(\mathbf{v})$. 
We estimate the mutual information between $X$ and $Y$ in the AWGN model using \eqref{eq:DIME_copula} and the generators described in Tab.~\ref{tab:generators}. Fig.~\ref{fig:i_x_y}a and Fig.~\ref{fig:i_x_y}b show the estimated mutual information for $d=1$ and $d=5$, respectively, and compare it with the closed-form capacity formula $I(X;Y) = d/2\log_2(1+\text{SNR})$. 

\begin{figure*}[t]
	\centering
	\includegraphics[width=\textwidth]{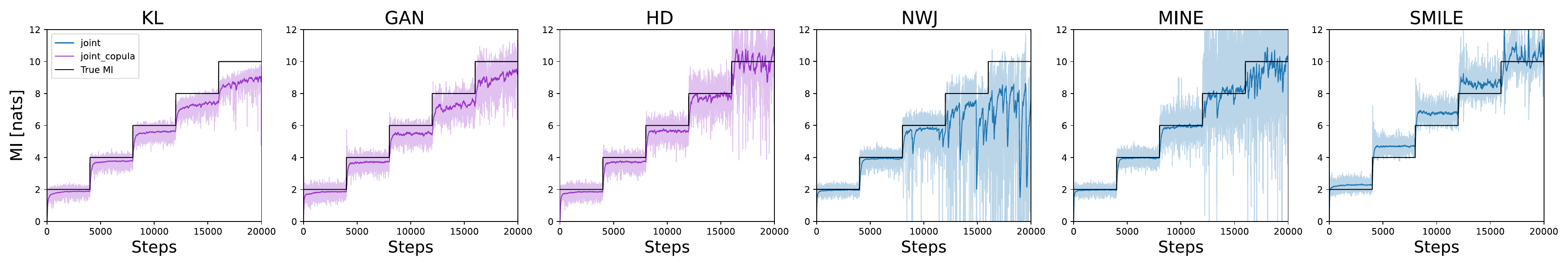}
	\caption{Estimated mutual information $I(X;Y)$ comparison for the asinh scenario between CODINE (in purple) implemented using the KL, GAN, and HD divergences, and NWJ \cite{Nguyen2010}, MINE \cite{Mine2018}, and SMILE \cite{Song2020} (in blue). }
	\label{fig:MI_comparison}
\end{figure*}

We also compare the usage of CODINE for mutual information estimation with state-of-the-art estimators \cite{Mine2018, Nguyen2010, Song2020} in Fig.~\ref{fig:MI_comparison}. We use the same neural architecture for all the estimators, referred to as \textit{joint} in the literature \cite{Mine2018}. We test the estimators over two-dimensional Gaussian random vectors to which we apply the inverse hyperbolic sine (asinh) mapping proposed in \cite{czyz2024beyond}, which shortens the tails of the Gaussian distribution. CODINE achieves comparable results with state-of-the-art mutual information estimators in terms of bias, and has significantly lower variance than NWJ and MINE. 
For MI estimation, the time requirements are the same as the density estimation task, as it only additionally require to get the samples from the product of marginals, which can be achieved in $O(1)$ with a shift-based derangement \cite{letizia2024Mutual}. 

\subsection{Data Generation via Gibbs Sampling}
\begin{figure*}
	\centering
	\includegraphics[scale=0.45]{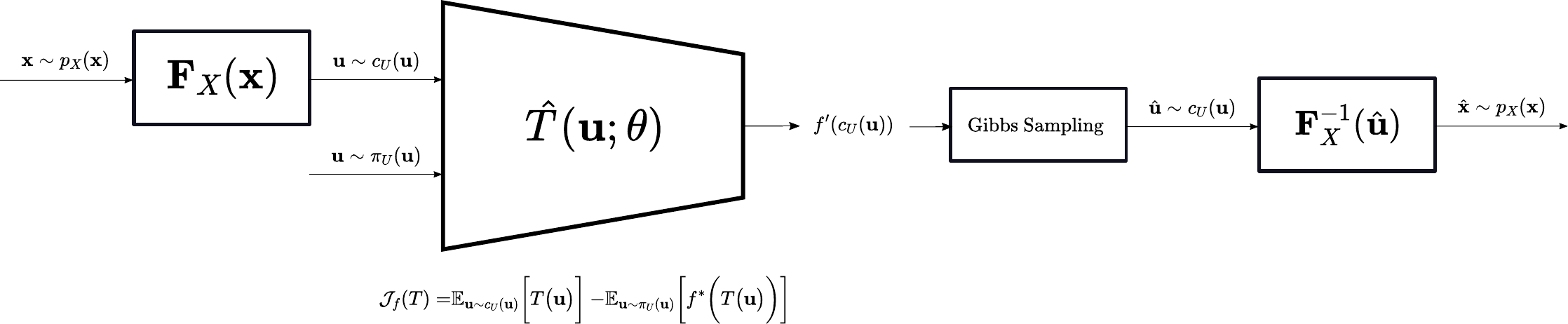}
	\caption{The analysis block utilizes CODINE to extract the copula density and the synthesis block utilizes Gibbs sampling to generate new data.}
	\label{fig:codine_generation}
\end{figure*}
As a second example of application, we generate $n$ new pseudo-observations $\{\hat{\mathbf{u}}^{(1)},\dots,\hat{\mathbf{u}}^{(n)}\}$ from $\hat{c}_U$ by deploying a Markov chain Monte Carlo (MCMC) algorithm. 
Validating the quality of the generated data provides an alternative path for assessing the copula estimate itself.

We propose to use Gibbs sampling to extract valid uniform realizations of the copula estimate. In particular, we start with an initial guess $\hat{\mathbf{u}}^{(0)}$ and produce next samples $\hat{\mathbf{u}}^{(i+1)}$ by sampling each component from univariate conditional densities $\hat{c}_U(u_j^{(i+1)}|u_1^{(i+1)},\dots,u_{j-1}^{(i+1)},u_{j+1}^{(i)},\dots,u_{d}^{(i)})$ for $j=1,\dots,d$.
It is clear that the generated data in the sample domain is obtained via inverse transform sampling through the estimated quantile functions $\hat{F}_{X_i}^{-1}$.
The proposed generation scheme is illustrated in Fig.~\ref{fig:codine_generation}.

\subsubsection{2D toy dataset}
\begin{figure}[t]
	\centering
	\includegraphics[scale=0.2]{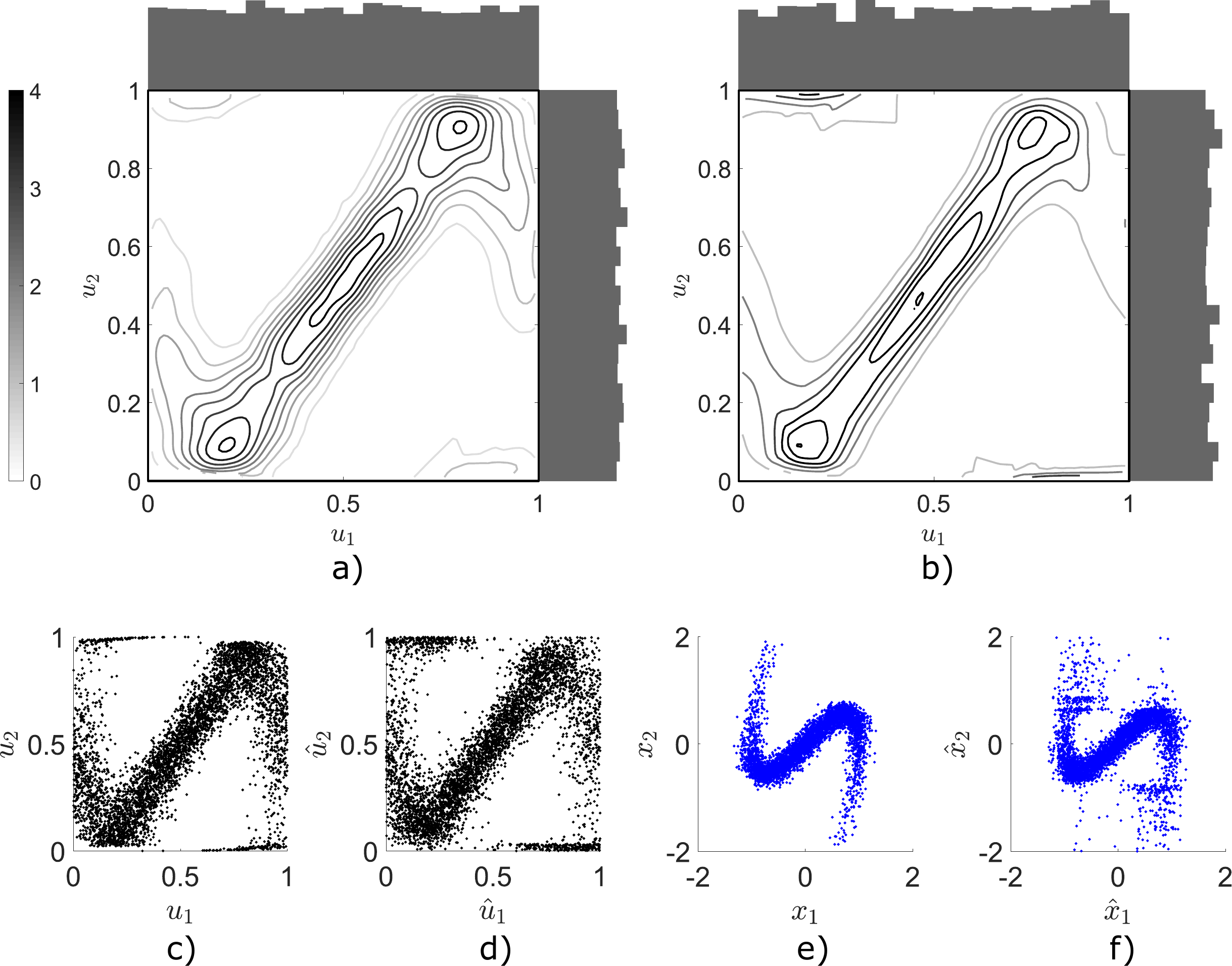}
	\caption{Toy example contour plot and marginal densities of a) ground-truth copula density $c_u$ obtained using kernel density estimation and b) copula density neural estimate $\hat{c}_u$. c) Pseudo-observations. d) Data generated in the uniform probability space via Gibbs sampling. e) Observations. f) Data generated in the sample domain via inverse transform sampling.}
	\label{fig:toy}
\end{figure}
Consider a bi-dimensional random variable whose realizations have form $\mathbf{x} = [\mathbf{x}_1, \mathbf{x}_2]$ and for which we want to generate new samples $\mathbf{\hat{x}} = [\mathbf{\hat{x}}_1, \mathbf{\hat{x}}_2]$. To force a non-linear statistical dependence structure, we define a toy example $\mathbf{x}$ as $
\mathbf{x} = [\sin(t), t\cos(t)] + \mathbf{n}$, 
where $t\sim \mathcal{N}(0,1)$ and $\mathbf{n}\sim \mathcal{N}(\mathbf{0},\sigma^2 \mathbb{I})$ with $\sigma = 0.1$. We use CODINE to estimate its copula density and sample from it via Gibbs sampling. Fig.~\ref{fig:toy} compares the copula density estimate obtained via kernel density estimation (Fig.~\ref{fig:toy}a) with the estimate obtained using CODINE (Fig.~\ref{fig:toy}b). It also shows the generated samples in the uniform (Fig.~\ref{fig:toy}d) and in the sample domain (Fig.~\ref{fig:toy}f). 
It is plausible that the Gibbs sampling mechanism produced some discrepancies between $\mathbf{x}$ and $\hat{\mathbf{x}}$.

\begin{figure}[t]
	\centering
	\includegraphics[scale=0.28]{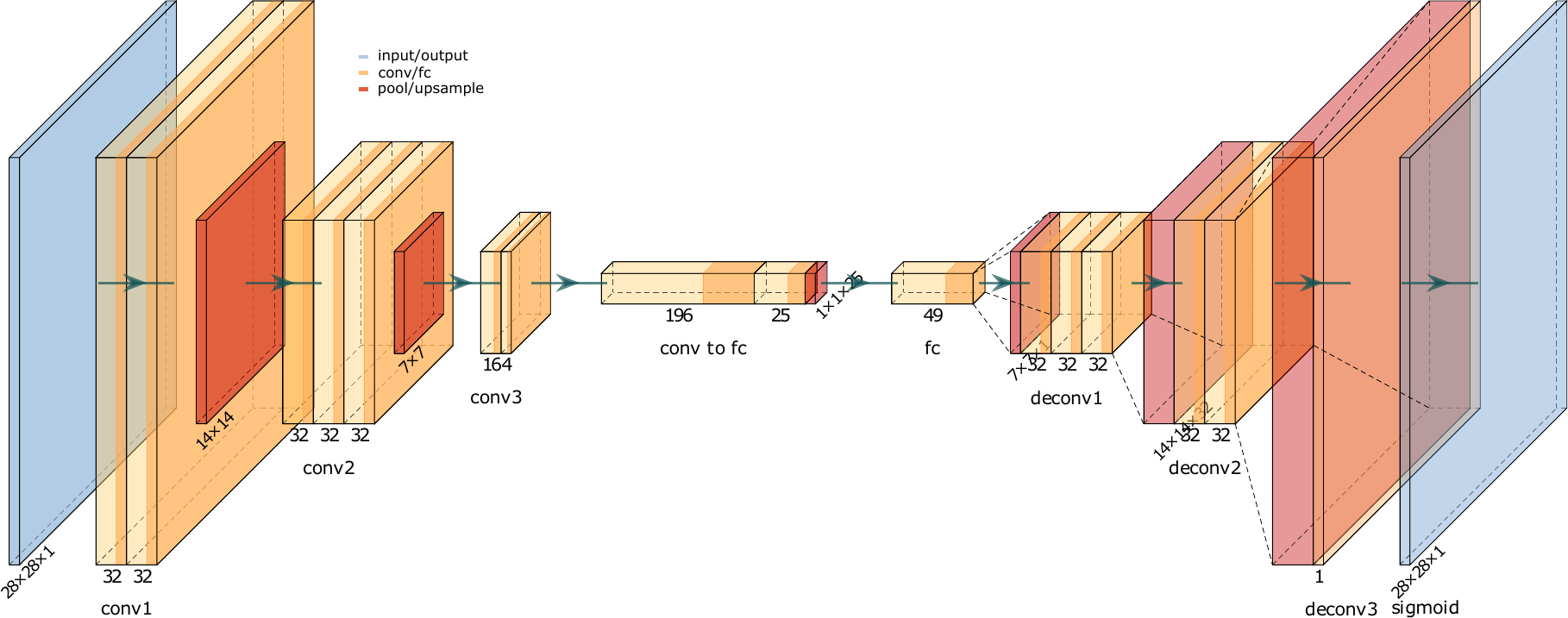}
	\caption{Autoencoder architecture used to learn latent vectors (e.g., $d=25$). The autoencoder is trained with binary cross-entropy for MNIST digits, and with MSE for FashionMNIST.}
	\label{fig:codine_architecture}
\end{figure}

\begin{figure}[t]
	\centering
	\includegraphics[scale=0.1]{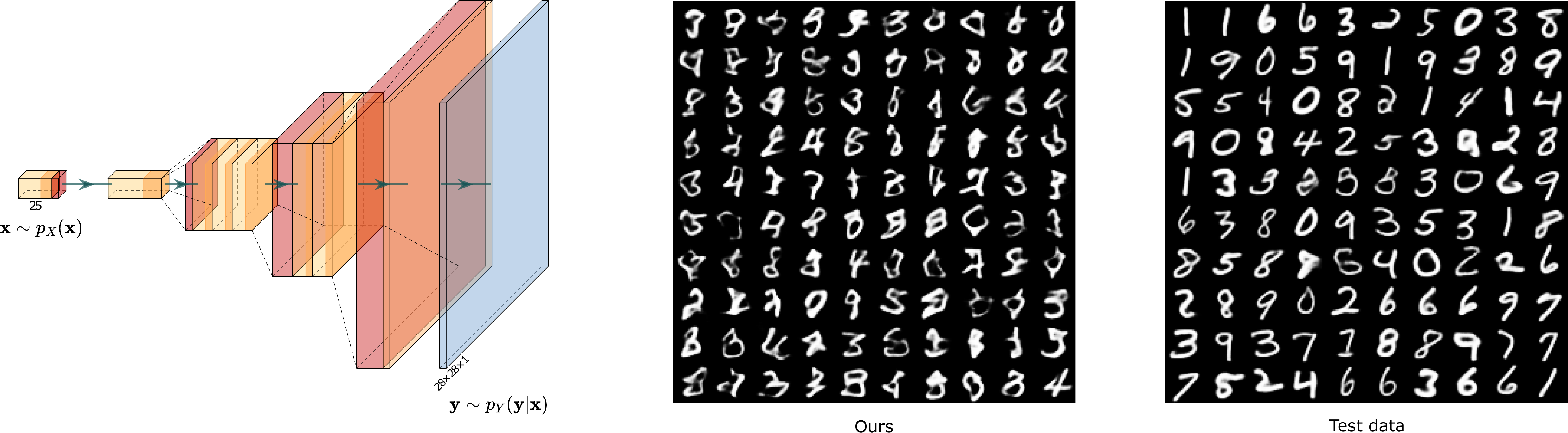}
	\caption{100 randomly selected digits obtained using CODINE to generate the latent vector of dimensions $d=25$ fed in the decoder. Comparison with test data.}
	\label{fig:codine_digits}
\end{figure}
\subsubsection{MNIST datasets}
\begin{figure}[t]
	\centering
	\includegraphics[width=\columnwidth]{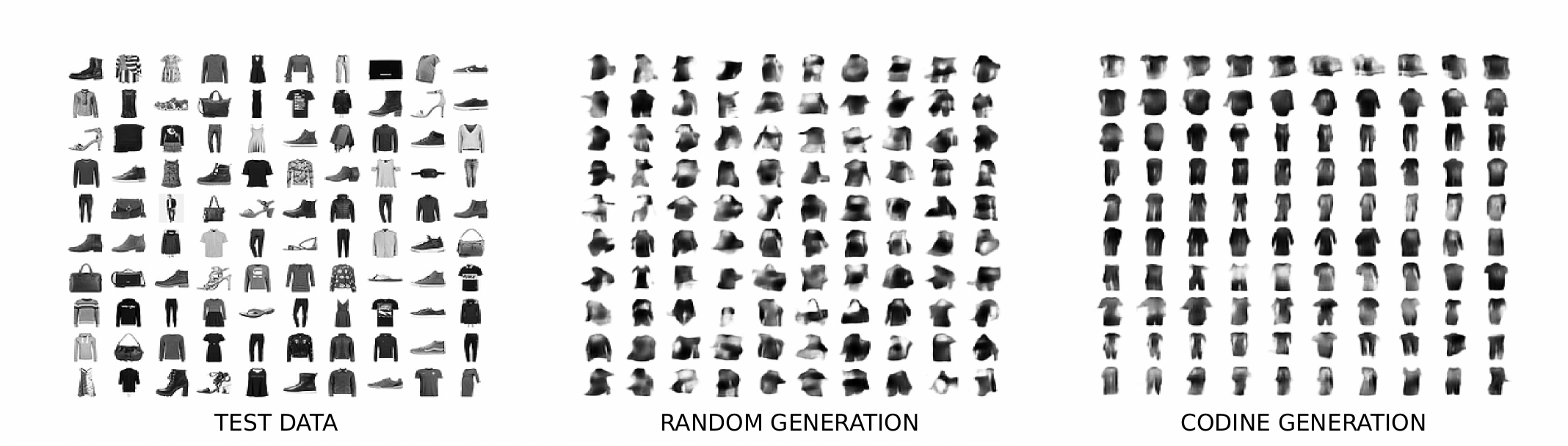}
	\caption{FashionMNIST comparison between test data and data generated using a trained decoder with latent space dimension $d=50$ starting from a random sampling of the latent space and from CODINE's copula estimate sampling.}
	\label{fig:codine_fashion_vs_random}
\end{figure}

\begin{figure}[t]
	\centering
	\includegraphics[width=\columnwidth]{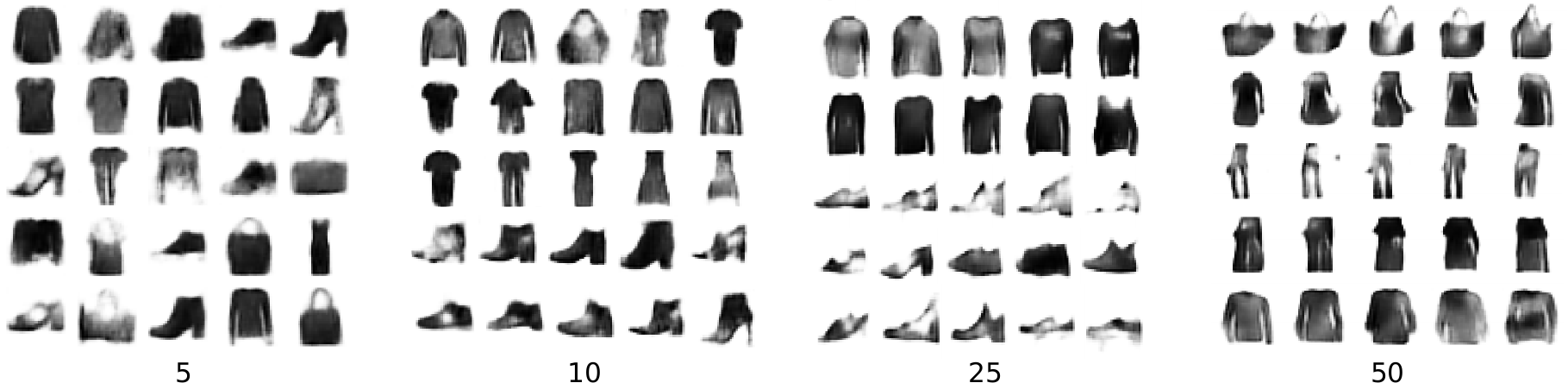}
	\caption{FashionMNIST generated images. They are obtained using CODINE to generate latent vectors of dimension $d=5, 10, 25, 50$ which are fed in the decoder.}
	\label{fig:codine_fashion}
\end{figure}

As more complex examples, we report the generation of the MNIST handwritten digits \cite{lecun1998gradient} and FashionMNIST \cite{xiao2017fashion} datasets of size $28 \times 28$. In particular, we study the copula density of the latent space obtained from an autoencoder. 
With these experiments, similarly to the experiments in \cite{tagasovska2019copulas, janke2021implicit}, we demonstrate that the approach works for high-dimensional data, and we want to emphasize that it is potentially scalable to higher dimensions. 
The idea is to train a convolutional neural network autoencoder to reconstruct the input images (see Fig. \ref{fig:codine_architecture}). During training, the autoencoder learns latent representations (via an encoder mapping) that can be analyzed and synthesized with CODINE. 
The analysis and synthesis strategies in Fig.~\ref{fig:codine_generation} comprise a first PIT block to project the latent representations into the uniform probability space. Then, a CODINE block learns the copula density which is used by Gibbs sampling to generate new uniform latent representations. An inverse transform sampling block maps data from the uniform to the sample space.
Once new latent samples are generated, it is possible to feed them into the pre-trained decoder and obtain new images, as illustrated in Fig.~\ref{fig:codine_digits} for the MNIST digits dataset. 
Although the CODINE block $T(\mathbf{u})$ is a simple shallow neural network, the generated digits visually resemble the original ones, meaning that our approach managed to, at least partially, estimate the density of the uniform latent representations. \\
To evaluate CODINE's ability to model and sample the latent space of the FashionMNIST dataset, we apply the same autoencoder and CODINE architectures used in our MNIST experiments. 
It is crucial to distinguish between effective image generation due to the decoder's power and valid latent distribution modeling. In fact, the "posterior collapse" problem, which causes many variational autoencoders with powerful decoders to have their latent spaces being ignored \cite{van2017neural}, could imply a good image generation without a proper learning of the latent space distribution. To isolate CODINE's contribution, we replace Gibbs sampling of the copula density learned by CODINE (in the pipeline previously described and depicted in Fig.~\ref{fig:codine_generation}) with a random generation of uniform samples. 
We compare the images generated from the trained decoder fed with samples from both methods in Fig.~\ref{fig:codine_fashion_vs_random}, demonstrating that CODINE's image generation is effective because: a) CODINE learns the copula of the latent distribution, and b) the decoder generates meaningful images from the latent space. \\
To evaluate CODINE's ability to model copulas of different dimensions, we vary the size of the latent space of the autoencoder. 
The comparison between the images generated with different latent space dimensions is showed in Fig.~\ref{fig:codine_fashion}, demonstrating the effectiveness of CODINE. 
While a higher latent space dimension yields more detailed images, it increases the complexity of accurately learning the copula density. To counteract the training difficulty, we notice that it is sufficient to increase the amount of training epochs of CODINE. For Fig.~\ref{fig:codine_fashion}, we train CODINE for 500, 500, 5k, and 20k epochs for $d$ equal to 5, 10, 25, and 50. 

In Appendix \ref{subsec:appendix_CODINEGAN}, we propose an additional method for data generation starting from CODINE's estimate of the copula density. We leave its analysis and implementation for future work. 

%% file: Conclusions.tex
\section{Conclusions}
\label{sec:conclusions}
This brief presented CODINE, a copula density neural estimator. It works by maximizing a variational lower bound on the $f$-divergence between two distributions defined on the uniform probability space, namely, the distribution of the pseudo-observations and the distribution of independent uniforms. 
The capability of CODINE in estimating any copula density further allows to tackle a wide variety of problems, such as mutual information estimation and data generation.

%% file: Appendix.tex
\section{Appendix}
\label{sec:appendix}

\subsection{Proof of Theorem \ref{theorem:Theorem1}}
\label{subsec:proof_thm1}
\begin{proof}
From the hypothesis, the $f$-divergence between $c_U$ and $\pi_U$ reads as follows
\begin{align}
D_f(c_U||\pi_U) = & \; \int_{\mathbb{R}^d}{\pi_U(\mathbf{u})f\biggl(\frac{c_U(\mathbf{u})}{\pi_U(\mathbf{u})}\biggr)\diff \mathbf{u}} \nonumber \\
= & \; \int_{[0,1]^d}{f\bigl(c_U(\mathbf{u})\bigr)\diff \mathbf{u}}.
\end{align}
Moreover, from Lemma 1 of \cite{Nguyen2010}, $D_f$ can be expressed in terms of its lower bound via Fenchel convex duality
\begin{equation}
\small
\label{eq:f_bound}
D_f(c_U||\pi_U) \geq \sup_{T\in \mathbb{R}} \biggl\{ \mathbb{E}_{\mathbf{u} \sim c_U(\mathbf{u})} \bigl[T(\mathbf{u})\bigr]-\mathbb{E}_{\mathbf{u}\sim \pi_U(\mathbf{u})}\bigl[f^*\bigl(T(\mathbf{u})\bigr)\bigr]\biggr\},
\end{equation}
where $T: [0,1]^d \to \mathbb{R}$ and $f^*$ is the Fenchel conjugate of $f$.
Since the equality in \eqref{eq:f_bound} is attained for $T(\mathbf{u})$ as
\begin{equation}
\hat{T}(\mathbf{u}) = f^{\prime} \bigl(c_U(\mathbf{u})\bigr),
\end{equation}
it is sufficient to find the function $\hat{T}(\mathbf{u})$ that maximizes the variational lower bound $\mathcal{J}_{f}(T)$.
Finally, by Fenchel duality it is also true that
$c_U(\mathbf{u}) = \bigl(f^{*}\bigr)^{\prime} \bigl(\hat{T}(\mathbf{u})\bigr)$.
\end{proof}

\subsection{Proof of Corollary \ref{cor:cor1}}
\label{subsec:proof_cor1}
\begin{proof}
From the hypothesis, the $f$-divergence between $p_X$ and $\pi_X$ reads as follows
\begin{align}
D_f(p_X||\pi_X) & = \int_{\mathbb{R}^d}{\prod_{i}{p_{X_i}(x_i)}f\biggl(\frac{p_X(\mathbf{x})}{\prod_{i}{p_{X_i}(x_i)}}\biggr)\diff \mathbf{x}} \nonumber \\
& = \int_{[0,1]^d}{f\bigl(c_U(\mathbf{u})\bigr)\diff \mathbf{u}},
\end{align}
where $c$ is the density of the copula obtained as in \eqref{SklarF}. The result then follows immediately from Theorem \ref{theorem:Theorem1}.
\end{proof}

\subsection{Generative Model Based on the Copula Estimate}
\label{subsec:appendix_CODINEGAN}
MCMC methods require increasing amount of time to sample from high-dimensional distributions. Thus, we study if it is possible to devise a neural sampling mechanism which exploit the copula density guidance. In particular, we exploit the maximum mean discrepancy (MMD) measure. Indeed, let $(\chi,d)$ be a nonempty compact metric space in which two copula densities, $c_U(\mathbf{u})$ and $c_V(\mathbf{v})$, are defined. Then, the MMD is defined as
\begin{equation}
\text{MMD}(\mathcal{G},c_U,c_V) := \sup_{g\in \mathcal{G}}\bigl\{\mathbb{E}_{\mathbf{u}\sim c_U(\mathbf{u})}[g(\mathbf{u})]-\mathbb{E}_{\mathbf{v}\sim c_V(\mathbf{v})}[g(\mathbf{v})]\bigr\},
\label{MMD}
\end{equation}
where $\mathcal{G}$ is a class of functions $g:\chi \rightarrow \mathbb{R}$. 
Since $c_U=c_V$ if and only if $\mathbb{E}_{\mathbf{u}\sim c_U(\mathbf{u})}[g(\mathbf{u})]=\mathbb{E}_{\mathbf{v}\sim c_V(\mathbf{v})}[g(\mathbf{v})]$ $\forall g\in \mathcal{G}$, MMD is a metric that measures the disparity between $c_U$ and $c_V$ (see \cite{FortetMMD}). If $g(\mathbf{u})=c_U(\mathbf{u})$ is a valid function, we can define a plausible loss function based on the MMD metric (and referred to as $\mathcal{J}_{\text{MMD}}(G_{\theta_G})$) as follows
\begin{equation}
\min_{\theta_G} \mathbb{E}_{\mathbf{u}\sim c_U(\mathbf{u})}[c_U(\mathbf{u})]-\mathbb{E}_{\mathbf{v}\sim \pi_V(\mathbf{v})}[c_U(G_{\theta_G}(\mathbf{v}))].
\label{eq:MMD-metric}
\end{equation}
Thus, given $n$ pseudo-observations $\{\mathbf{u}^{(1)},\dots,\mathbf{u}^{(n)}\}$ for which we have built and estimated the underlined $c_U(\mathbf{u})$, it is possible to design a neural network architecture, the generator $G$, which maps independent uniforms with distribution $\pi_V$ into uniforms with distribution $c_{V,\theta_G}$. The guidance provided by $c_U(\mathbf{u})$ helps minimizing the discrepancy between the two copulas when the optimization is performed over $\theta_G$. The optimal generator resulting from the solution of \eqref{eq:MMD-metric} synthesizes new pseudo-observations $\hat{\mathbf{u}} = G(\mathbf{v};\theta_G^*)$.

To verify if $c_U$ is a properly defined function, it is useful to notice that the Wasserstein metric, and in particular the Kantorovich Rubinstein duality links with the MMD in \eqref{MMD} for a class of functions $\mathcal{G}$ that are $K$-Lipschitz continuous
\begin{equation}
\label{eq:WGAN}
W(c_U,c_V) = \frac{1}{K}\sup_{||h||_L\leq K}\biggl[ \mathbb{E}_{\mathbf{u}\sim c_U(\mathbf{u})}[h(\mathbf{u})]-\mathbb{E}_{\mathbf{v}\sim c_V(\mathbf{v})}[h(\mathbf{v}))\biggr],
\end{equation}
where $||\cdot||_L$ is the $L$-th norm.
Under such conditions, \eqref{eq:MMD-metric} can be interpreted as the generator loss function of a Wasserstein-GAN \cite{WGAN} where the optimum discriminator $h$ is supposed to be known and corresponds to the learnt copula density $c_U$. The proposed idea lies in between two established approaches. The first one, generative moment matching networks GMMNs \cite{GMMNs}, assume $\mathcal{G}$ as the reproducing kernel Hilbert space where $g$ is a kernel $k \in \mathcal{H}$ and the supremum in \eqref{MMD} is thus attained. Such MMD-based optimizes only over the generator's parameters but does not produce expressive generators, mainly because of the restriction imposed by the kernel structure. The second, instead, requires to learn both the generator and the discriminator, the latter in order to reach the supremum in \eqref{eq:WGAN}. However, enforcing the Lipschitz constraint is not trivial and the alternation between generator and discriminator training suffers from the usual instability and slow convergence problems \cite{Goodfellow2014}. Even if the copula-based approach does not claim optimality, it possesses two desirable properties: compared to kernel-based methods, it uses a more powerful and appropriate discriminator, the copula density itself. Moreover, the fact that $c_U$ is obtained from a prior analysis renders the generator learning process uncoupled from the discriminator's one.